\pdfoutput=1

\documentclass[11pt]{article}

\usepackage[final]{acl}

\usepackage{times}
\usepackage{latexsym}

\usepackage[T1]{fontenc}

\usepackage[utf8]{inputenc}

\usepackage{microtype}

\usepackage{inconsolata}

\usepackage{graphicx}
\usepackage[noalg]{stroop}
\usepackage{bbm}
\usepackage[]{todonotes}
\newcommand{\langpair}[2]{\texttt{#1-#2}}
\newcommand\knnmt{$k$-NN MT}
\newcommand\knn{$k$-NN}

\title{Angular Dispersion Accelerates  $k$-Nearest Neighbors Machine Translation}

\author{Evgeniia Tokarchuk \\
   University of Amsterdam\\
  \texttt{evgeniia@tokarch.uk} \\\And
  Sergey Troshin \\
  University of Amsterdam \\
  \texttt{s.troshin@uva.nl} \\\And 
  Vlad Niculae\\
  University of Amsterdam \\
  \texttt{v.niculae@uva.nl} \\}

\begin{document}
\maketitle

\begin{abstract}
Augmenting neural machine translation with external memory at decoding time, 
in the form of
$k$-nearest neighbors machine translation (\knnmt{}), 
is a well-established strategy for increasing translation performance.
\knnmt{} retrieves a set of tokens that occurred in the most similar contexts recorded in a prepared data store, using hidden state representations of translation contexts as vector lookup keys.
One of the main disadvantages of this method is the high computational cost and memory requirements. Since an exhaustive search is not feasible in large data stores, practitioners 
commonly use approximate \knn{} 
lookup, yet even such algorithms are a bottleneck. 
In contrast to research directions seeking to accelerate \knnmt{} by reducing
data store size or the number of lookup calls,
we pursue an orthogonal direction based on the performance properties of 
approximate \knn{} lookup data structures. In particular, we propose
to encourage
angular dispersion of the neural hidden representations of contexts.
We show that improving dispersion leads to better balance in the
retrieval data structures, accelerating retrieval
and slightly improving translations.
\end{abstract}

\section{Introduction}
\begin{figure}
    \centering
    \includegraphics[width=0.9\linewidth]{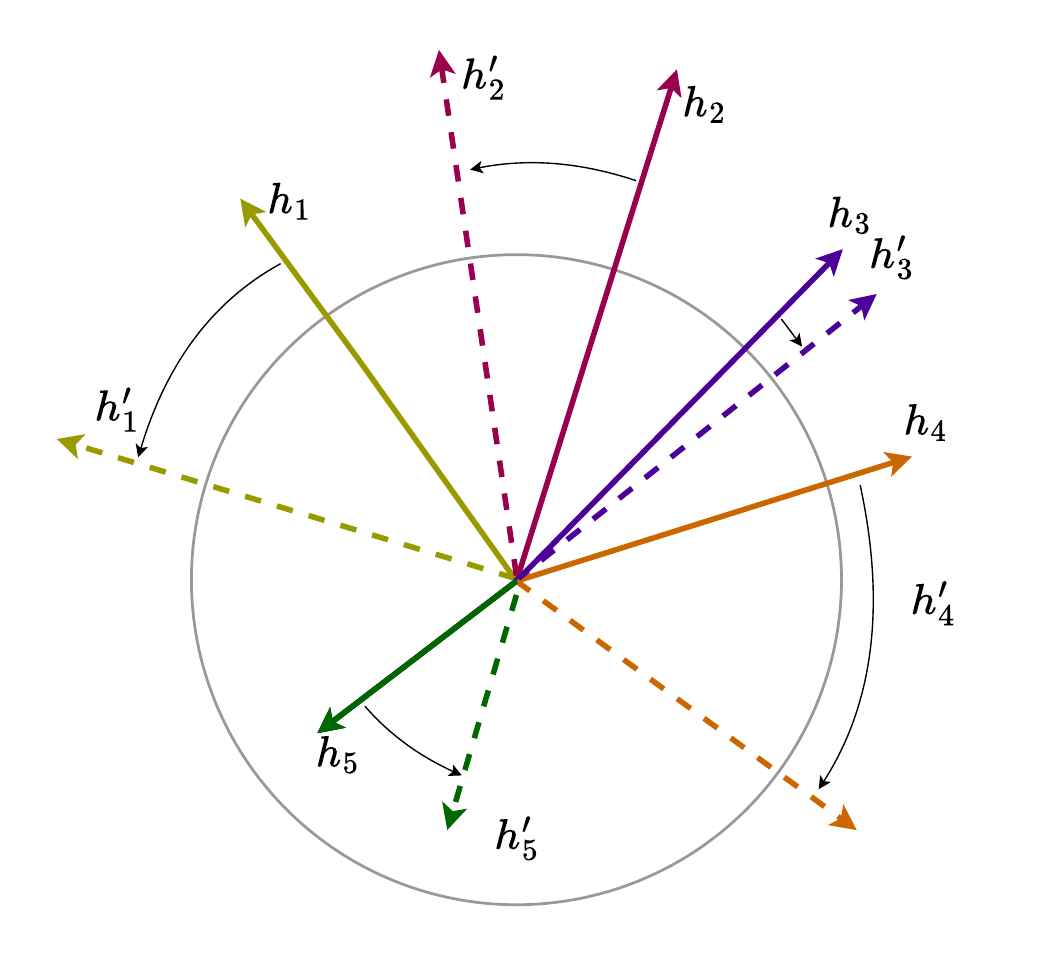}
    \caption{\label{fig:ang-disp} Angular dispersion applied to a set of vectors: dispersion spreads out the angles of the vectors, without changing their length (norm). By increasing angular dispersion of key-value data stores, we observe substantial speed-ups at the retrieval time. Here, solid lines represent vectors before dispersion, and dashed lines after dispersion.
    }
\end{figure}

$k$-Nearest Neighbors Machine Translation (\knnmt) is a promising training-free approach demonstrated to
improve the translation quality of neural MT in both in- and out-of-domain setups \citep{khandelwal2021nearest, meng-etal-2022-fast,dai2023simple}. However, performance improvement comes with a high decoding cost because \knn{} queries require nearest-neighbor lookups in a usually large \textit{data store}:
a key-value store where tokens in the training dataset are indexed by continuous key vectors (given by the hidden states of the neural MT system when processing the context).

With modern neural networks, keys tend to be high-dimensional ($\geq 512$), and, given the number of tokens in a data store, \knn{} search requires substantial time and space.
Thus, in order to search this large data collection, approximate \knn{} \citep{johnson2019billion} is dominant in practice \citep{khandelwal2021nearest,martins-etal-2023-empirical,gao-etal-2024-efficient}. However, even approximate \knn{} methods can not achieve a decoding speed of the baseline models out-of-the-box.

Many works focus on the efficiency of retrieval either by data store compression \citep{he-etal-2021-efficient,dai2023simple,wang-etal-2022-efficient, martins-etal-2022-efficient, zhu-etal-2023-knowledge} or by adaptive retrieval at the decoding time \citep{martins-etal-2022-efficient, martins-etal-2022-chunk, gao-etal-2024-efficient}. 
Key representations have, perhaps surprisingly,
received less attention, even though their geometric properties
play an important role in retrieval.
\citet{wang-etal-2022-learning-decoupled} argue that key vectors are not specific enough for fine-grained retrieval and suggest a contrastive learning approach to decouple representations. Similarly, \citet{wang-etal-2022-efficient} show that there is a large overlap between data store keys, 
and that the overlap is not related to word frequency; they
suggest a method based on contrastive learning that can mitigate the overlap.

Since modern data structures 
that allow billion-scale distance-based \knn{} search \citep{johnson2019billion}
rely heavily on space partitioning, the distribution of keys affects not only the quality of the retrieved samples but also plays a significant role in retrieval speed due to the cluster imbalance \citep{Tavenard_Jegou_Amsaleg_2011}. \citet{gao2018representation} and \citet{ethayarajh-2019-contextual} draw attention to the clumping of the continuous representations of Transformer-based \citep{Vaswani-trafo} models, \ie, they show that model representations tend to occupy only a small subspace and semantically different words cluster together.
Motivated by the work of \citet{tokarchuk2025distancelearningdispersedembeddings}, we study distributional properties of the keys in \knnmt{}, focusing on \textit{angular dispersion} (\Cref{fig:ang-disp}) . In particular, we show that higher angular dispersion of the key vectors leads to up to 5 times more efficient \knn{} lookups in large (>10M) data stores. We also emphasize that angular dispersion results in an overall slight increase of the translation performance for both \knnmt{} with in-domain and out-of-domain data stores.

\section{Background}
\subsection{k-NN Machine Translation}
\label{sec:knnmt}
We follow the 
\knnmt{} model proposed by \citet{khandelwal2021nearest}, in which a neural MT system is augmented with a non-parametric $k$-nearest neighbors classifier at the token level. 
Given a set of pairs of source and target sentences $(\mathcal{X},\mathcal{Y})$, and a well-trained encoder-decoder model with parameters $\theta$, 
let $h_\theta(x, y_{<i}) \in \bbR^d$ denote a hidden state representation of the context needed for translating the $i$-th output token \footnote{%
Often the last hidden layer in a transformer decoder.
}.
Given a key-value data store, we associate the key $h_\theta(x, y_{<i})$ with the target $y_i$,
for all prefixes of all pairs in the parallel dataset.
This results in a data store of key-value pairs \(D=\{(h_j, y_j)\}_{j=1}^N\), where $N$ is the total number of contexts used (on the order of millions).
Given some new query state $h$ and a distance function $d$, the $k$-nearest neighbor set $N_k(h)$ is a subset of $D$ containing exactly $k$ pairs such that if $(h', y') \in N_k(h)$ and $(h'', y'') \in D \setminus N_k(h)$ then $d(h, h') \leq d(h, h'')$ .
The $k$-nearest neighbors set can be used to induce a probabilistic classifier from $h$ onto the vocabulary, essentially by counting each neighboring state as a ``vote'' toward that word, weighted by the distance: 
\begin{multline}
    p_{\text{$k$-NN}}\left(y|h\right) \propto \\ \sum_{(h_j,y_j) \in N_k(h)} \llbracket y=y_j \rrbracket \exp \left(\frac{-\|h - h_j\|^2}{T}\right),
\end{multline}
where $\llbracket \cdot \rrbracket$ is the Iverson bracket, equal to 1 or 0 depending on the truth value of its argument, $T>0$ is a temperature parameter adjusting the flatness of the induced distribution, and $\propto$ denotes equality up to a normalization constant.
When translating from the neural MT system,
we interpolate its own predictive distribution with the one of the $k$NN classifier with weight $0 \leq \lambda \leq 1$:
\begin{multline}
    p(y_i|x, y_{<i}) := (1-\lambda)p_{\text{model}}(y_{i}| x,y_{<i}) \\ + \lambda p_{\text{$k$-NN}} \left(y_i | h_\theta(x, y_{<i})\right).
\end{multline}

\subsection{Approximate k-NN search}
\label{sec:ivfpq}
The efficiency bottleneck of \knnmt{} is the nearest-neighbor search. 
For this reason, the search is typically not performed exactly; rather,
it is common to employ
\textit{approximate} nearest-neighbor search,
which enables larger data stores up to billions of tokens \citep{johnson2019billion}. We follow the line of \knnmt{} research \citep{khandelwal2021nearest, zheng-etal-2021-adaptive, meng-etal-2022-fast, wang-etal-2022-efficient, martins-etal-2022-chunk, deguchi-etal-2023-subset, deguchi2023knnseqefficientextensibleknnmt} and use the inverted file index with product quantization (IVFPQ),
through the implementation in the \texttt{faiss} library \citep{johnson2019billion},
which, with GPU acceleration, is state-of-the-art in terms of accuracy and speed for this application.

\begin{figure}
    \centering
    \includegraphics[width=0.95\linewidth]{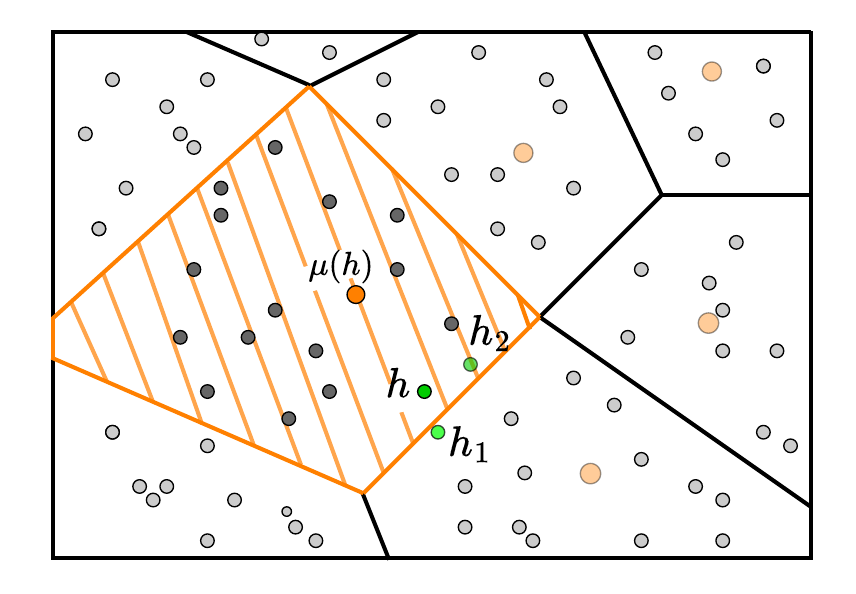}
    \caption{An illustration of the IVF index with a single probe. A query vector $h$ is mapped to the nearest centroid $\mu(h)$. The search is then performed within the Voronoi cell of $\mu(h)$ (orange area), and then  $h_2$ will be returned as the approximate nearest neighbor of $h$. With larger nprobes, it is possible to retrieve $h_1$, the closest neighbor of $h$.}
    \label{fig:IVF_demo}
\end{figure}

\paragraph{Inverted file index (IVF).} IVF first splits the keys $h$ into clusters by using $k$-means clustering algorithm \citep{K-Means} by learning $K$ centroids $\mu=[\mu_1, \ldots, \mu_K]$. Given the set of learned centroids, any vector $h\in\mathbb{R}^d$ is  associated with its nearest centroid as $\mu(h)=\argmin_{\mu'} \|h-\mu'\|$. The search vector space is then divided into Voronoi cells  (see \Cref{fig:IVF_demo}) and the data store is split into clusters $D_{\mu_i} = \{(h_j, y_j) \,|\, \mu(h_j)=\mu_i, (h_j, y_j) \in D \}$. In practice, searching within a single IVF cluster is suboptimal, and several nearest IVF clusters (probes) are searched \citep{johnson2019billion}. Given a query vector, we retrieve a shortlist of nearest neighbor candidates from the top-$n$ (nprobes) nearest Voronoi cells: $\text{IVF}(h) = \{(h_j, y_j) \,|\, \mu(h_j)\in K_{\textit{IVF}}(h), (h_j, y_j) \in D\}$, where $K_{\textit{IVF}}(h)$ denotes the set of nearest centroids given a query vector, namely probes. %

\paragraph{Product quantization (PQ).} Product quantization \citep{product_q_jegou_2011} compresses the $d$-dimensional key vectors for fine-grained search. While IVF is the part of the data structure that we most target in our work, we describe the PQ stage for completeness.  PQ splits the vector space into a Cartesian product of $M$ subspaces, each $d/m$--dimensional: $h = [\tilde{h}_1, \ldots, \tilde{h}_M]$, and quantizes each subspace separately. For each subspace, we use $k$-means to learn $L$ $d/M$-dimensional codewords per subspace: $C = [c_i]_{i=1}^{L}$. To quantize a key, one finds the closest codeword ids within each subspace, namely for a given subspace we search for the closest codeword using the Euclidean distance between a codeword and $\tilde{h}$ (the sub-vector of $h$): $\text{PQ}(\tilde{h}) = %
\argmin_i
\| c_i - \tilde{h} \|$.  A standard practice is to represent each key vector using $8$ bits.

\paragraph{IVFPQ.} This strategy combines the inverted file index and product quantization. To store the keys, PQ compresses the residual key representations within an IVF cluster, where the residual key representation is a key vector minus the associated IVF centroid. Given a query vector, IVF first maps the query to a Voronoi cell of the nearest IVF centroid. Then, PQ performs a fine-grained search within the Voronoi cell using the stored quantized key representations.

The efficiency and accuracy of lookups in the 
IVFPQ data structure depends on the quality of the clustering stages, which depends in turn on the geometric distribution of the key vectors \citep{Tavenard_Jegou_Amsaleg_2011}. Quantifying the properties of a point cloud that make it well-suited for IVFPQ lookup remains an open question. We propose to analyze the geometry of the \emph{directions} of the key vectors, and, in particular, their \emph{angular dispersion}.

\subsection{Angular Dispersion}

Directions in $d$-dimensional space can be represented as
points on the sphere $\bbS_d \subset \bbR^d$.
Unlike the entirety of $\bbR^d$, the sphere is compact and has many computationally attractive properties that allow us to quantify and optimize angular dispersion.
Given two directions $s, s' \in \bbS_d$, their Euclidean dot product $\DP{s}{s'} \in [-1,1]$
corresponds to the cosine of the angle between them, and thus
the angle is $\arccos\DP{s}{s'} \in [0, \pi].$
A well-dispersed configuration $S \in (\bbS_d)^n$
is spread out uniformly over the entire sphere surface. One way to quantify dispersion
is by the angle between the closest points:
\begin{equation}
d_{\text{min}}(S)\coloneqq \min_{s \neq s' \in S}
\arccos\DP{s}{s'}.
\end{equation}
An alternative and complementary measure of angular dispersion is
spherical variance \citep{alma999052553502466,mardia1975statistics}:
\begin{equation}
    \label{eq:svar}
    \operatorname{svar}(S) = 1-\left\|\frac{1}{|S|}\sum_{s \in S} s \right\|.
\end{equation}
Spherical variance close to $1$ indicates high concentration of directions, as the average will be close to the sphere surface; a value close to $0$ does not necessarily imply dispersion but at least a more symmetrical configuration of the directions: for any $x \in \bbS_d$, the four-point configuration $(x,x,-x,-x)$ has both minimum angle and spherical variance zero.

Finding an optimally-dispersed configuration is known as the Tammes problem \citep{tammes-1930}, and it is generally not exactly tractable.
Instead, it is common to define functions to optimize in order to encourage dispersion
\citep{pmlr-v119-wang20k,wang2020mmadispersion,liu2018learning,tokarchuk2025distancelearningdispersedembeddings}. 
A wide class of such functions are based on pairwise similarities or \emph{kernels}. For example, the \emph{minimum hyperspherical energy} objective is
\begin{equation}
    \label{eq:mhe}
    R_{\text{MHE}}(S) = \frac{1}{|S|(|S|-1)}\sum_{s \neq s' \in S} k(s, s').
\end{equation}
where $k(s,s')$ is a kernel such as perhaps the Gaussian kernel $k(s,s') = \exp(\DP{s}{s'}/\sigma)$.

\emph{Sliced dispersion}
\citep{tokarchuk2025distancelearningdispersedembeddings,bonet2023spherical}
is an efficient alternative which avoids the quadratic complexity by making use of the fact that optimal dispersion is trivial on a circle (on $\bbS_2 \subset \bbR^2$). In this special case, any perfectly-dispersed configuration is made up of equidistant angles and given any input set of angles, the nearest dispersed configuration can be efficiently found. 
Let the sum of angles between a configuration $S$ and the nearest optimally-dispersed configuration on a circle be $\delta(S)$.
If $d=2$ we can exactly minimize $\delta(S)$. To extend to higher dimensions, we will invoke projections onto great circles.
On a sphere $\bbS_d$, the great circles correspond to pairs of orthogonal directions \(C(\bbS_d) \coloneqq \{(p, q): p \in \bbS_d, q \in \bbS_d, \DP{p}{q}=0\}\).
If we denote by $S_{pq}$ the projection of a configuration of directions $S$ onto the great circle $(p,q)$,
sliced dispersion optimizes
\begin{equation}
R_\text{sliced}(S) \coloneqq \bbE_{p,q} \left[ \delta(S_{pq})\right],
\end{equation}
where the expectation is over the uniform distribution on $C(\bbS_d)$.
In words, this objective minimizes the expected distance to an optimally-sliced configuration along any great circle.
Preliminary experiments in \Cref{app:comparison-reg} show that sliced dispersion regularizer converges faster in terms of spherical variance, so we use it for all further experiments.

\section{Key Representation with Dispersion}\label{sec:dispersion}
Dispersion on the sphere is well-established in the literature as discussed in \Cref{sec:dispersion}, and it is common to apply dispersion of the angles for learning hyperspherical representations, \ie, magnitudes of the vectors are discarded. However, previous works show that the norm of the keys, in fact, might contain important information \citep{gao-etal-2024-efficient}. To avoid discarding norms, contrary to other works, we apply dispersion only to the angles of the model's outputs while keeping their magnitudes intact and do gradient updates in Euclidean space. 
\Cref{fig:ang-disp} shows how angular dispersion operates in $2D$. 

\paragraph{Optimization for dispersion.}In the context of \knnmt{}, we also have to avoid a mismatch between the data store representation and the model's output to keep the decoding step consistent. Therefore, we cannot apply dispersion on the data store alone. We fine-tune a small portion of a pretrained model to optimize for the dispersion of the model's outputs before they are saved in a data store. We estimate dispersion over all possible keys using the mini-batches.

Given a dataset of paired (parallel) sentences $(\mathcal{X}, \mathcal{Y})$ as in \cref{sec:knnmt}, 
we employ the standard log-likelihood machine translation loss \footnote{In practice, we additionally use label smoothing regularization \citep{label_smooth} with $\epsilon=0.1$.} \wrt the trainable model parameters $\theta$:
\begin{equation}
    \mathcal{L_{\text{MT}}}(\theta) \coloneqq -
    \sum_{(x,y) } %
    \sum_{i=1}^{|y|} \log p(y_i|x,y_{<i}).
\end{equation}
In addition, given a dispersion regularizer $R$ that acts on the sphere $\bbS_{d}$, we define a dispersion loss on the directions of the hidden states. Concretely, writing
\(H\coloneqq\left\{h(x, y_{<i}) :
(x,y) \in \mathcal{X}\times\mathcal{Y},
i \in |y|\right\} \subset \bbR^d \) and 
\(\bar{H} = \{h/\|h\|: h \in H\} \subset \bbS_d\), we have

\begin{equation*}
    \mathcal{L_{\text{Disp}}}(\theta) \coloneqq
    R(\bar{H}),
\end{equation*}
and we optimize their weighted sum

\begin{equation}\label{eq:finetuning-loss}
    \mathcal{L}(\theta) \coloneqq
    \mathcal{L}_{\text{MT}}(\theta) + 
    \gamma\mathcal{L}_{\text{Disp}}(\theta),
\end{equation}
estimating its stochastic gradients on minibatches.

Any directional dispersion regularizer can play the part of $R$; 
we use sliced dispersion as motivated in~\Cref{sec:dispersion}.

\begin{figure}[ht]
    \centering
    \includegraphics[width=.6\linewidth]{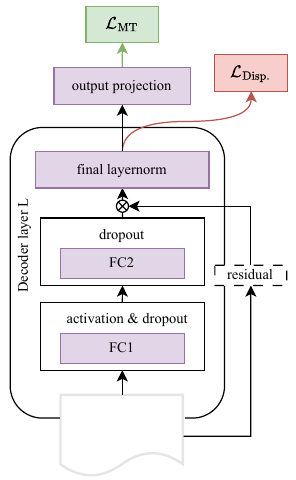 }
    \caption{\label{fig:ft-for-dispersion}Fine-tuning of the NMT model for dispersion of outputs. Blocks in purple are trained, and the rest of the parameters are frozen.}
\end{figure}

During the entire dispersion process, most of the network parameters are kept frozen. We only fine-tune the weights of the last two feed-forward layers, layer normalization, and output projection of the final decoder block. Hence, we do not introduce any new model parameters and utilize the existing architecture. Schematically, \Cref{fig:ft-for-dispersion} shows how we update the model representations during training.

\paragraph{Key representation.}In kNN-MT, the input to the final Transformer decoder fully-connected (FC) block is usually used as the key representation for the datastore \citep{khandelwal2021nearest}. In our setup (\Cref{fig:ft-for-dispersion}), following \citet{martins-etal-2023-empirical}, we instead use the \textit{decoder output} as the key, and keep this choice consistent across experiments. Preliminary tests showed a small advantage for the conventional choice (input to the final FC decoder block) in vanilla kNN-MT. However, when fine-tuning, we found that adjusting the output projection is essential for good performance. This requires fine-tuning all weights up to the output, making the conventional approach less practical as it involves more parameters. How to fine-tune for dispersion more efficiently remains an open question.

\section{Experimental Setup}
\subsection{Synthetic Data}\label{sec:synthdata}
To more directly control the effect of dispersion, we generate a family of synthetic data stores.
Specifically, we sample 10M random vectors of dimensionality 128 from the mixture of 5 power spherical distributions \citep{decao2020power}. The scale parameter $\kappa$ for the power spherical distribution defines the \textit{concentration} of the data points, where higher $\kappa$ means higher concentration. We build a data store using the keys generated with $\kappa=\{1,10,50, 100,1000\}$ with 2048 centroids while picking a random length for each key in the $[1,100]$ range.

\subsection{MT Datasets and Evaluation}
We provide empirical results on two different languages with in-domain and out-of-domain data stores. Below is the description of the datasets and evaluation criteria.
\begin{itemize}
    \item WMT16 \langpair{ro}{en} news translation dataset\footnote{https://www.statmt.org/wmt16/} with 612k training samples. We apply \texttt{sentencepiece}\footnote{\url{https://github.com/google/sentencepiece}} tokenization \citep{kudo-2018-subword}.
    \item Multi-domain \langpair{en}{de} dataset \citep{aharoni2020unsupervised}. We prepare data similar to \citet{khandelwal2021nearest}, applying BPE \citep{sennrich-etal-2016-neural} and data filtering based on sentence length\footnote{\url{https://github.com/urvashik/knnmt/blob/master/examples/translation/prepare-domadapt.sh}{}}. The number of training samples varies by domain: \texttt{Medical} (206K), \texttt{Law} (450K), \texttt{IT} (180K), and \texttt{Koran} (15K).
    \item WMT19 \langpair{en}{de} dataset provided by  HuggingFace\footnote{\url{https://huggingface.co/datasets/wmt/wmt19}}. We use this dataset only for fine-tuning \langpair{de}{en} translation model. It contains 34M filtered \langpair{de}{en} news data.
\end{itemize}
 We evaluate the translation accuracy on the best checkpoint according to the validation BLEU score using SacreBLEU~\citep{papineni-etal-2002-bleu,post-2018-call} and COMET~\citep{rei-etal-2020-comet} on \texttt{newstest2016} for \langpair{ro}{en} and test sets of four domains (\texttt{Medical}, \texttt{Law}, \texttt{IT} and \texttt{Koran}) for \langpair{de}{en}. We do all experiments using \texttt{fairseq} framework~\citep{ott2019fairseq}.

\subsection{Models}
We compare translation quality and decoding speed
across the following models: 
\begin{itemize}
    \item \textbf{Baseline (base).}  We train a transformer-base model \citep{Vaswani-trafo} for \langpair{ro}{en}, with additional linear projection of the decoder outputs to dimensionality 128. We deliberately choose this dimensionality since the preliminary results show a performance increase compared to a model with output dimensionality 512 (see \Cref{app:add-results}). For \langpair{de}{en}, we use the pre-trained model \citep{ng-etal-2019-facebook} with the dimensionality of the decoder output 1024.
    \item \textbf{Baseline fine-tuned with the dispersion objective (base-D).} For both \langpair{ro}{en} and \langpair{de}{en} we fine-tune the baseline model as shown in \Cref{fig:ft-for-dispersion}. All layers, except the last two fully connected layers, layer normalization of the last decoder block, and an output projection, are frozen during fine-tuning. For \langpair{de}{en} fine-tuning, we use WMT 19 training set, and for \langpair{ro}{en} we use WMT 16 training set. We fine-tuned all models with 5k optimization steps. We choose $\gamma$ equal to $1$ for all experiments and use the sliced dispersion regularizer with one great circle per batch.
    \item \textbf{Vanilla \knnmt{} with (\knn{}-D) and without (\knn{}) dispersion.} 
    These models are obtained without additional training, by combining the aforementioned models (base and base-D)
    with \knn{} interpolation at test time \citep{khandelwal2021nearest}.
    The data store is built from the model checkpoint with best development BLEU.
\end{itemize} 
More details of the model training and inference can be found in \Cref{app:training-det}.

\subsection{Data Store Construction and Search}
To build a data store from a NMT model, we extract the hidden states from the last decoder feed-forward layer $h^{(L)}$. 
As described in \Cref{sec:ivfpq},
we use \texttt{faiss} with IVFPQ index \citep{johnson2019billion} and train the index on 1M random samples\footnote{For data stores with less than $1M$ elements, we use all elements.}. For all models, we use 2048 IVF centroids and retrieve according to the squared Euclidean distance.

\section{Results}

\subsection{Synthetic Results}
\begin{figure}
    \centering
    \includegraphics[width=\linewidth]{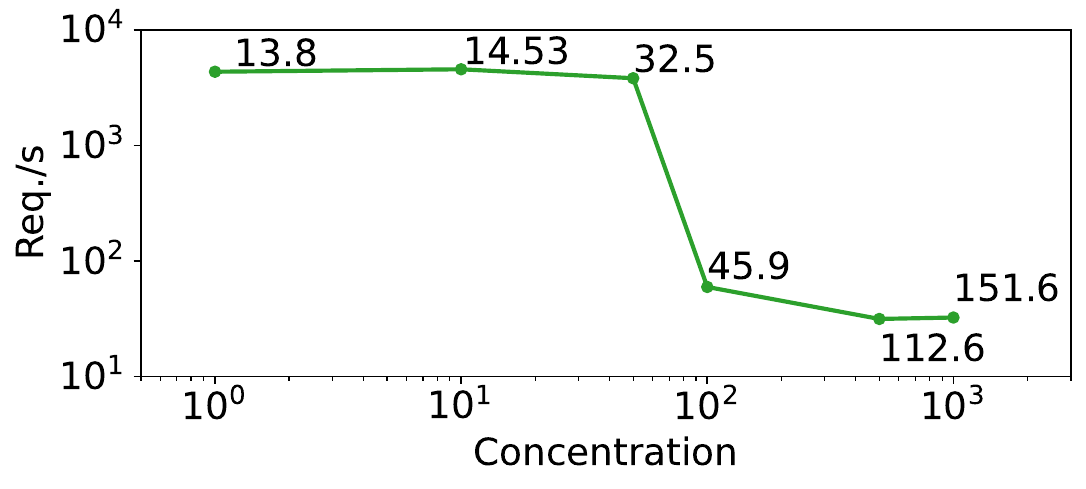}
    \caption{\label{fig:synthetic-exp} The number of requests per second (req./s) for the data stores generated with different concentration parameters $\kappa$. Higher concentration means lower dispersion. We also show the imbalance factor for each data store above the req./s value. Notably, increasing or decreasing dispersion past certain points does not influence speed much, but a certain critical range of concentration makes a substantial difference when crossed.} 
\end{figure}

We first verify our hypothesis in a controlled experiment. 
Using the collection of synthetic data stores indexed by concentration (\cref{sec:synthdata}),
we measure the time to retrieve $k=8$ nearest neighbors for 10K random queries with batch size 10 and nprobe $32$. 
\Cref{fig:synthetic-exp} shows that the data store 
lookup speed (measured as the number of queries per second) correlates negatively with the dispersion. 

\subsection{In-domain Results}
\label{sec:roenres}
\begin{figure}
    \centering
    \includegraphics[width=\linewidth]{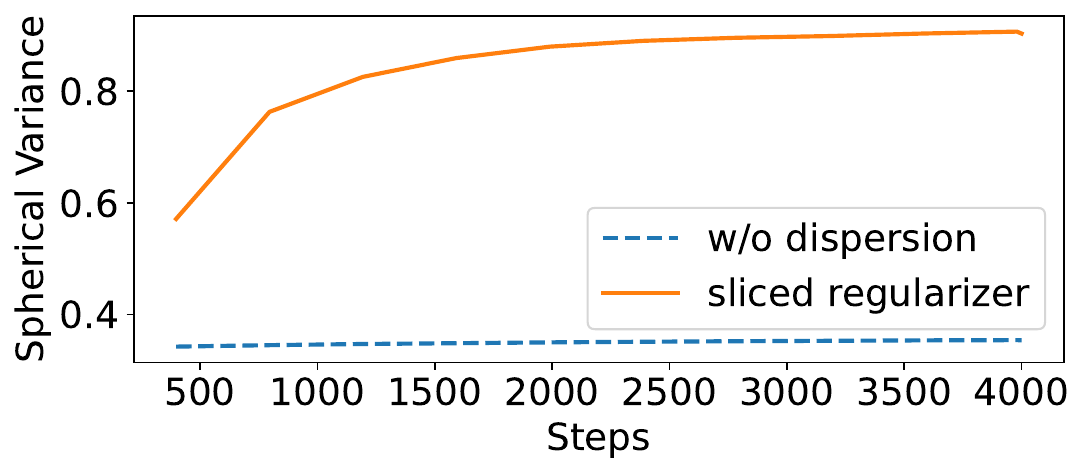}
    \caption{\label{fig:ro-en-dynamic} Fine-tuning dynamics of \langpair{ro}{en} models in terms of spherical variance. A step is a gradient update on the effective batch size described in \Cref{app:training-det}.}
    \end{figure}

First, we analyze fine-tuning dynamics of the model with sliced regularizer described in \Cref{sec:dispersion} and without dispersion regularizer.
We only report spherical variance, as the number of keys is highly large, and minimum distance has quadratic complexity.
\Cref{fig:ro-en-dynamic} shows that fine-tuning the model with the sliced regularizer increases the spherical variance value to a large extent, compared to the fine-tuning without the dispersion regularizer.

\Cref{tab:ro-en-128} shows translation results for base and base-D models, both with and without \knn{}. We also vary the number of centroids to probe between 32 and 8. The total amount of keys stored in the data store is equal to 21M. Alongside the translation scores, we report the number of tokens processed in one second (tok/s).
\begin{table}[ht]
    \centering
    \small
    \setlength{\tabcolsep}{3pt} %
    \begin{tabular}{lrrrr}
    \toprule
    \textbf{model} & \textbf{\#pr.} & $\textbf{BLEU}_{(\uparrow)}$  & $\textbf{COMET}_{(\uparrow)}$  & $\textbf{tok/s}_{(\uparrow)}$\\
    \midrule
        base & - &31.5 & 78.95 & 75  \\ 
        base-D &  - & 31.7 & 78.96 & 79 \\ \midrule  
        \knn{}  & 32& 32.4 & 79.89 &  12 \\ 
        \knn{}  & 8& 32.2 & 79.69 & 28 \\ 
        \knn{}-D & 32& 32.6  & 79.91 & 53 \\ 
        \knn{}-D & 8& \textbf{32.6}  & \textbf{79.93} & \textbf{63} \\ 
    \bottomrule
    \end{tabular}
    
    \caption{\label{tab:ro-en-128}\langpair{ro}{en} translation scores on \texttt{newstest16} test set. tok/s calculated as a median over three runs.}
\end{table}

\knnmt{} decoding with dispersion (base-D) performs as well as the base \knnmt{} model while being about 5 times faster. 
The speedup agrees with the synthetic results,
and so we attribute it to the properties of the IVFPQ index.
To verify this hypothesis, we conduct an analysis of the data store properties in \Cref{sec:geom-analysis} and establish a connection to the NMT model performance.

We provide additional results with embeddings dimensionality equal to 512 in \Cref{app:add-results}.

\subsection{Domain Adaptation}\label{sec:domadapt-res}
Following the line of \knnmt{} works \citep{khandelwal2021nearest,gao-etal-2024-efficient,meng-etal-2022-fast}, we provide results for five different domains for \langpair{de}{en}, comparing the models with and without dispersion. As shown in \Cref{tab:de-en results}, a model with dispersion can achieve slightly faster data store lookup and overall better performance on all domains. The main difference between the two setups is that 
all \langpair{de}{en} keys have larger dimension ($1024$ instead of $128$), and the data store size, except for \texttt{Law}, is much smaller than for \langpair{ro}{en}. 

Our experiments also show that domain adaptation \knnmt{} is robust against decreasing the number of clusters to probe, while without dispersion, we can see a performance drop both in BLEU and COMET. The most pronounced lookup speed improvements can be seen on the \texttt{Law} domain since it has the largest \knn{} data store out of the four domains. Note that we fine-tuned for dispersion using only WMT 19 training data and did not use any domain-specific data. We measure average tok/s over 10 random subsets, each consisting of 200 sentences drawn uniformly with replacement from the test set. We report the median measurement.

\paragraph{Translation analysis.}
While studying the difference in the system outputs using \texttt{compare-mt} \citep{neubig-compare-mt}, we do not see many systematic patterns that can plausibly explain the performance gain of the $k$-NN MT-D model. We find that sentence length is predicted more accurately in the dispersed model and sentence-level quality increases for short sentences. We leave further investigation to future work.

\begin{table*}[ht]
    \centering
    \small
    \setlength{\tabcolsep}{4pt} %
    \begin{tabular}{lrrccrccrccrcc}
    \toprule
        \multirow{2}{*}{\textbf{model}} & \multirow{2}{*}{\textbf{\#pr.}} & \multicolumn{3}{c}{\textbf{\texttt{Medical}} (5.7M)} & \multicolumn{3}{c}{\textbf{\texttt{Law}} (18.4M)} & \multicolumn{3}{c}{\textbf{\texttt{IT}} (3.1M)} & \multicolumn{3}{c}{\textbf{\texttt{Koran}} (0.5M)}  \\ 
        &                   & BLEU & COMET            & tok/s & BLEU & COMET & tok/s & BLEU & COMET & tok/s & BLEU & COMET & tok/s  \\ \midrule
        base        & -     & 40.4 & 83.05            & 91    & 45.8 & 85.24 & 90 & 37.6 & 82.07  & 70 & 16.9 & 72.48 & 89  \\
        base-D      & -     & 40.5 & 83.30            & 86    & 46.0 & 85.37 & 92 & 38.3 & 82.45  & 82 & 17.1 & 72.58 & 76  \\ \midrule
        \knnmt{}    & 32    & 55.0 & 84.41            & 55    & 61.0 & 87.03 & 17 & 45.1 & 83.13  & 51 & 20.8 & 72.38 & 62  \\
        \knnmt{}    & 8     & 54.6 & 84.37            & 63    & 60.7 & 86.87 & 36 & 44.6 & 83.12  & 55 & 20.1 & 71.44 & 63  \\
        \knnmt{}-D  & 32    & \textbf{55.2} & \textbf{84.54} & 62    & \textbf{61.5} & \textbf{87.07} & 41 & \textbf{45.8} & \textbf{83.54} & 55 & \textbf{21.4} & \textbf{72.58} & 62\\ 
        \knnmt{}-D  & 8     & 55.1 & 84.47            & 66    & 61.3 & 87.01 & 47 & 45.8 & 83.49 & 58 & 21.3 & 72.27 & 63 \\
        \bottomrule
    \end{tabular}%
    \caption{\label{tab:de-en results} Translation quality (BLEU/COMET scores) on different domains for \langpair{de}{en} alongside the tok/s decoding speed. The \#npr column shows the number of probes used during approximate $k$-NN retrieval (\cref{sec:ivfpq}).}
\end{table*}

\section{Analysis and Data Store Geometric~Properties}
\label{sec:geom-analysis}
Empirically, dispersion has a positive effect in terms of lookup speed for the synthetic and \knnmt{} experiments. To further explore how dispersion affects the IVFPQ search, we measure how it affects the properties of the IVF clusters.
\paragraph{Dimensionality.}
Intuitively, clumping occurs more often when dimensionality is relatively small. Previous studies on dispersion ~\citep{tokarchuk2025distancelearningdispersedembeddings} and our experiments discussed in \Cref{sec:roenres,sec:domadapt-res} show that higher dimensionality alleviates clumping problem to some extent. However, increasing dimensionality does not necessarily leads to better performance (\eg, \langpair{ro}{en} model with output dimensionality 128 performs overall better than the same model with output dimensionality 512, as per \Cref{tab:ro-en-128} and \Cref{tab:roen512}), but it often incurs additional computation and memory costs. While having larger dimensionality showed to be useful in large-scale setups~\citep{}, with \knnmt{} such scaling is non-trivial given the size of resulting data store. However, there is a performance and speed gain even with higher dimensionality as per \Cref{tab:de-en results}. Therefore, dispersion might be a good and relatively cheap tool that helps balancing clusters in the data store, and as a result improve retrieval speed with no additional handcrafted optimizations. We do believe that our method can be successfully applied in scenarios where large amount of points have to be distributed in high-dimensional space.

\begin{figure}
    \centering
    \includegraphics[width=\linewidth]{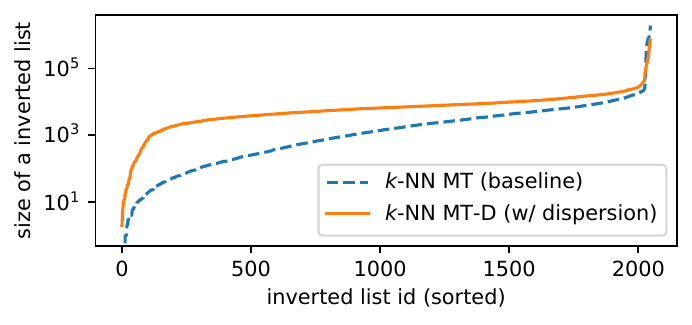}
    \caption{Dispersion tends to reduce the variance of IVF cluster sizes.}
    \label{fig:ivf_cluster_sizes}
\end{figure}

\paragraph{Distribution of IVF cluster sizes.} We investigate whether dispersion improves the distribution of cluster sizes, which is known to have a positive effect on the retrieval speed and leads to a more predictable retrieval time \citep{Tavenard_Jegou_Amsaleg_2011}. In the \Cref{fig:ivf_cluster_sizes}, we compare the distributions of the IVF cluster sizes for the \knnmt{} and the \knnmt{}-D data stores. 
To numerically quantify the effect, we report the imbalance factor \citep{Tavenard_Jegou_Amsaleg_2011}. Given a partitioning of a data store elements into $K$ clusters with sizes $N_i$, and the total size $N=\sum_{i=1}^K N_i$, the imbalance factor is:
\begin{equation}
\text{IF} = K \sum_{i=1}^K \left(\frac{N_i}{N}\right)^2.
\end{equation}
We observe that dispersion improves the imbalance factor (IF), as observed from \Cref{tab:ro-en-128-clustering} and \Cref{fig:synthetic-exp}.

\paragraph{Number of clusters to probe.}
Next, we are asking whether dispersion improves the quantization accuracy such that we need fewer IVF probes to obtain the same quality.

To complement the analysis above, we also propose the expected number of probes (ENP) metric that aims to reflect the number of probes that is enough to obtain high-quality retrieval. First, given a vector $h$, we rank the IVF centroids $\mu =[\mu_i]_{i=1}^K$, by the distance from the $h$: $R_h(\mu_i)\in\{1,\ldots,K\}$. Next, given $k$
 nearest neighbors $[h_i]_{i=1}^k \in K(h)$, and their corresponding IVF centroids ranks $\mu_h=[\mu(h_i)]_{i=1}^k$, we estimate the ENP metric:
 \begin{equation}
 \text{ENP} = \mathbb{E}_{h\sim \mathcal{D}} \max_{h' \in K(h)} R_h(\mu(h')).
 \end{equation}
 ENP takes values from $1$ to $K$, where $1$ means $1$ is the perfect score (meaning we need $1$ probe to find $k$-NNs). We use the approximate search with $32$ probes to estimate the reference $[h_i]_{i=1}^k$.

 For the \langpair{ro}{en} data stores, the ENP metric for \knnmt{} is $6.296 \pm 0.667$, while for \knnmt{}-D the ENP is $3.290 \pm 0.215$; for \langpair{de}{en} (\texttt{Law}) data stores, we observe a marginal improvement in ENP ($1.59 \pm 0.056$ vs $1.5 \pm 0.049$). The improvements in ENP are more pronounced for the \langpair{ro}{en} data stores. We attribute this to the difference in key dimensionality and the amount of training data used.

\paragraph{Clustering algorithm evaluation.}
Further, we are asking whether dispersion improves other clustering metrics. Namely, we look at the commonly used clustering metrics, namely \textit{homogeneity}, \textit{completeness}, and \textit{$v$-measure} \citep{rosenberg-hirschberg-2007-v} implemented via scikit-learn \citep{scikit-learn}. 
Motivated by the precision, recall, and $f$-measure \citep{vanRijsbergen1979}, homogeneity, completeness, and $v$-measure estimate the quality of a clustering given keys and classes (labels), where we use the next tokens $v$ from the vocabulary $V$. \textit{Homogeneity} measures if all of its clusters contain only data points that are members of a single class; \textit{completeness} measures if all the data points that are members of a given class are elements of the same cluster; and \textit{$v$-measure} is the harmonic mean between homogeneity and completeness. For the definitions, we refer the reader to \cref{app:clustering_analysis}.

As we show in \Cref{tab:ro-en-128-clustering}, the clustering performance improves for the dispersed model. In particular, homogeneity and $v$-measure improve for both \langpair{ro}{en} and \langpair{de}{en} data stores by a large margin, while completeness improves for the \langpair{ro}{en} data store.

\begin{table}[ht]
    \centering
    \small
    \setlength{\tabcolsep}{2pt} %
    \begin{tabular}{lcccc}
    \toprule
    \multirow{2}{*}{\textbf{metric}} & \multicolumn{2}{c}{\langpair{ro}{en}} & \multicolumn{2}{c}{\langpair{de}{en}} \\
    & w/o disp. & w/ disp. & w/o disp. & w/ disp. \\ \midrule
    homogeneity ($\uparrow$) & 51.54 & 71.98 & 67.56 & 74.06   \\
    completeness ($\uparrow$) &  44.41 & 73.90 & 70.27 & 70.11  \\
    $v$-measure ($\uparrow$) & 61.39 & 70.16 & 68.89 & 72.03  \\ 
    IF          ($\downarrow$) & 68.34 & 11.12 & 27.73 & 7.79 \\

    \bottomrule
    \end{tabular}
    \caption{\label{tab:ro-en-128-clustering} 
    Clustering metrics for in-domain \langpair{ro}{en} \knnmt{} and \texttt{Law} \langpair{de}{en} \knnmt{} data stores with and without dispersion. Dispersion improves homogeneity and $v$-measure for both \langpair{ro}{en} and \langpair{de}{en} data stores.
    }
\end{table}

\paragraph{Symmetry of IVF keys.}
One of the ways to quantify the balance of the representation space is to measure how symmetrical the space is, \ie, we are asking whether the dispersion helps to increase the symmetry of the key space. We measure the central symmetry as the Euclidean norm of the mean vector over the data store. Namely, if the norm of the mean vector is low, the vector representations must be well-balanced relative to the origin. Given a set of keys $h$, sampled from the data store $\mathcal{D}$, we estimate $\norm{\mathbb{E}[h]}$. For the \langpair{ro}{en} data stores, $k$-NN MT  $\norm{\mathbb{E}[h]} = 68.35$, and $k$-NN MT-D $\norm{\mathbb{E}[h]} = 11.12$; for \langpair{de}{en} (\texttt{Law}) data stores,  $k$-NN MT $\norm{\mathbb{E}[h]} = 6.03$, and $k$-NN MT-D $\norm{\mathbb{E}[h]} = 2.91$, which indicates that dispersion improves the symmetry of the vector representations\footnote{Note that symmetry around zero implies zero mean, but zero mean does not necessarily imply symmetry.}.

\subsection{Sensitivity to Dispersion Weight}
In order to verify that speed-up and performance gains are the positive effect of dispersion rather than additional fine-tuning, we vary the value of $\gamma$ in \Cref{eq:finetuning-loss} where $\gamma=0$ is equivalent to fine-tuning with only cross-entropy loss. Results in \Cref{tab:de-en-gamma-sens} that positive values of gamma lead to improvements in both speed and translation quality, while being not sensitive to the specific value of $\gamma$.
\begin{table}[ht]
    \centering
    \small
    \setlength{\tabcolsep}{4pt} %
    \begin{tabular}{lcrcrcr}
    \toprule
        \multirow{2}{*}{\textbf{$\gamma$}} & \multicolumn{2}{c}{\textbf{\texttt{Medical}} (5.7M)} & \multicolumn{2}{c}{\textbf{\texttt{Law}} (18.4M)} & \multicolumn{2}{c}{\textbf{\texttt{IT}} (3.1M)}  \\ 
         & BLEU   & tok/s & BLEU & tok/s & BLEU  & tok/s \\ \midrule
        -    & 53.8  & 53.2 & 60.4 & 16.2 & 42.8 & 48.5   \\
        0 & 53.9  &  46.7    & 60.5 & 27.5 & 43.0 &  48.2  \\
        1 & 54.2  & 51.2 &  60.7  & 40.4 & 43.0 & 53.0\\ 
        100 &  53.7 & 53.7 & 61.4 & 46.6 & 43.3 &  55.0  \\
        \bottomrule
    \end{tabular}%
    \caption{\label{tab:de-en-gamma-sens} BLEU scores on different domains for \langpair{de}{en} development sets alongside the tok/s decoding speed with various values of $\gamma$.
    }
\end{table}

\section{Related Work}
In the literature, there are two main lines of work addressing the decoding speed of \knnmt{}.
One of them is related to the creation of the data store and includes methods such as pruning and dimensionality reduction. Another direction of the \knnmt{} improvements lies in the decoding-time improvement, including caching and adaptive retrieval. \citet{he-etal-2021-efficient} discuss various pruning strategies (including greedy merge pruning) and dimensionality reduction in the context of \knn{} language models. \citet{martins-etal-2022-efficient} apply similar techniques in the context of NMT domain adaptation with \knn{}.
\citet{zhu-etal-2023-knowledge} propose to prune data stores based on a local correctness metric, which is defined as the maximum number of correct predictions in the neighborhood of a specific key. \citet{wang-etal-2022-efficient} and \citet{meng-etal-2022-fast} build a source-side small data store to avoid searching across all possible target contexts. \citet{dai2023simple} also use a small data store and perform sentence-based retrieval. They incorporate results into the NMT model using an adapter network. %

\citet{Haokui-2022-conn-compress} also show that there are redundant features in the keys and that a compression network can be learned for efficient search, while \citet{zhu-etal-2023-ink} use adapter and alignment loss to learn low-dimensional keys representation.

To improve the decoding speed at the decoding time, \citet{martins-etal-2022-chunk} introduced chunk-based retrieval that retrieves multiple tokens at once instead of one token at a time. \citet{martins-etal-2022-efficient} use a caching mechanism in decoding time and a simple decision mechanism for skipping retrieval from the data store based on a predicted interpolation value $\lambda$ rather than a static $\lambda$ as in the original \knnmt{} model \cite{khandelwal2021nearest}. Similar in nature, \citet{gao-etal-2024-efficient} propose to use a classifier to decide when to skip retrieval from the \knn{} data store and introduce a timestep aware \knnmt{} threshold $\lambda$.

In contrast, the study of the key representation properties is limited. \citet{wang-etal-2022-efficient} analyze the distributions of the keys and show that even unrelated words can be seen in the same clusters, which can negatively affect the distance-based retrieval. They train adapter network with contrastive loss to promote more separable representations and introduce compression. Similarly \citet{wang-etal-2022-learning-decoupled} show that context representation from the NMT model is suboptimal for the \knnmt{} retrieval and more fine-grained representation can be obtained via training small adapter network with a contrastive loss approach. In our work, we show that a similar goal can be achieved with a simple regularizer without the need for negative samples. Also, we show that we can achieve speed up even without compression. 

Orthogonal to the speed-up of the \knnmt{} is the design choice of the \knn{} search index itself that affects retrieval speed greatly. In our work, we have focused on IVF-PQ since it is an index that that is well established and validated in the \knnmt{} literature and methodology \citep{khandelwal2021nearest,martins-etal-2022-efficient,gao-etal-2024-efficient}. Our method has direct impact on improving IVF lookup of the index rather then improving search over quantized vectors (PQ). Using such methods as Optimized Product Quantization \citep[OPQ,][]{Ge-OPQ-2014} might further improve search speed.

\section{Conclusion} In this work, we show that angular dispersion of the data store benefits the retrieval speed and performance in \knnmt{} for both in-domain and domain adaptation scenarios. Our analysis of cluster properties and synthetic data stores further indicates that dispersion of the keys balances the data store clustering, accelerating the approximate search, especially for larger data store sizes. Our findings indicate that fine-tuning with dispersion can be a first efficient \knnmt{} component. Since there are no changes in architecture or \knn{} data store construction, we hypothesize that other efficient methods that rely on distance metrics can be easily applied on top, which we leave for future work. %

\section*{Limitations}
\paragraph{Limited study of dimensionality effect.} Dispersion and dimensionality have a tight connection with each other. Specifically, larger dimensionality naturally allows for large dispersion but induces greater costs for storage and retrieval. However, in this work, we focus on three different choices of dimensionality: 128, 512 and 1024. The dimensionality of modern LLMs is typically on the larger side, and it remains a question to what extent we can see the benefits of dispersion in application to language modeling, particularly at a larger scale.
\paragraph{Search Metric.} Similar to previous works, we use squared Euclidean distance for the \knn{} retrieval. \citet{wang-etal-2022-learning-decoupled} specified in their work that using the same metric as in fine-tuning leads to better performance. We use angular dispersion, and it is possible that resorting to angular-based distances, such as cosine similarity, during pretraining and fine-tuning will lead to better performance of the dispersed models.
\paragraph{Data store Index Design}
Our analysis relies on a specific index design with a fixed number of IVF centroids. Among all parameters, the number of centroids plays a key role in determining data store lookup speed. In addition, optimized product quantization can further accelerate index search. In this work, however, we do not explore modifications to the data store index, and leave this direction for future investigation.

\section*{Acknowledgments}
We thank Caio Corro and the UvA Language Technology Lab, especially Yang Meng, Maya Nachesa, Di Wu and Melika Davoodzadeh for discussion and
feedback on the manuscript. This work is supported by the Dutch Research Council (NWO) via VI.Veni.212.228. The authors also thank SURF (www.surf.nl) for the support in using the National Supercomputer Snellius.

\bibliography{custom}

\appendix

\section{Regularizers}
\label{app:comparison-reg}
\begin{figure}[ht]
    \centering
    \includegraphics[width=\linewidth]{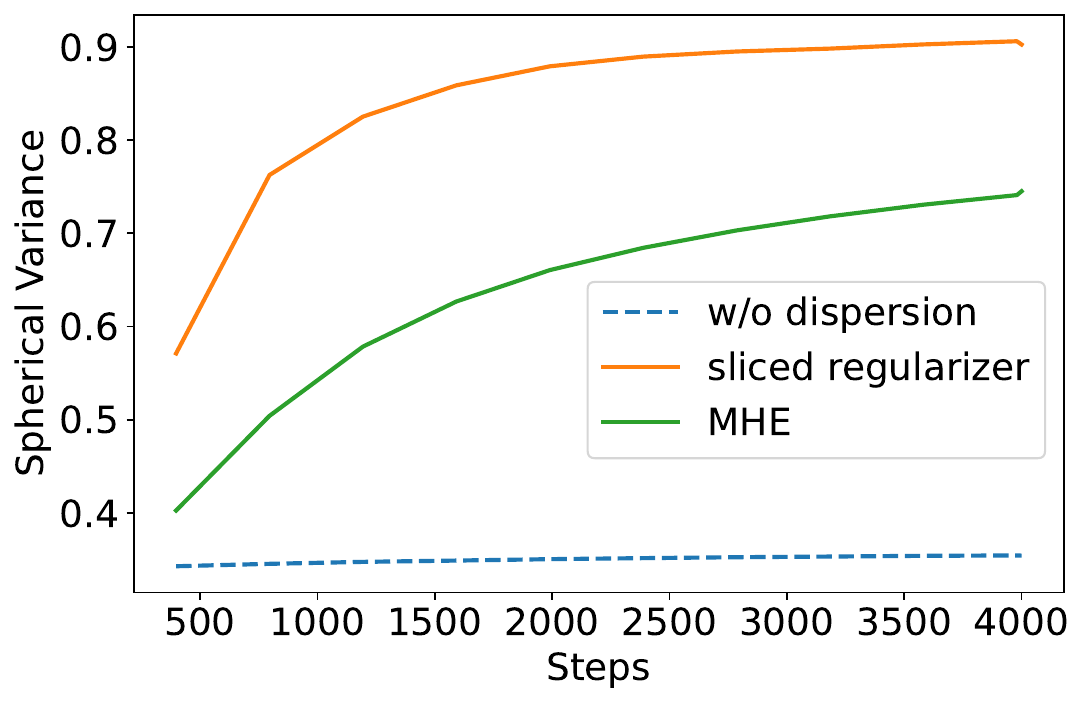}
    \caption{\label{fig:spherical-v-withmhe}Comparison of the MHE and sliced regularizer.}
\end{figure}

We compare the behavior of sliced and MHE regularizers (see \Cref{sec:dispersion}) during the \langpair{ro}{en} model fine-tuning. \Cref{fig:spherical-v-withmhe} shows batched spherical variance for both regularizers on the training data. We can see that sliced requires fewer updates for convergence, and the overall spherical variance is higher. Therefore, we use the sliced regularizer for all experiments. 

\section{Training and Inference Details}
\label{app:training-det}
\paragraph{Fine-tuning.} We fine-tune base models with cross-entropy loss, label smoothing equal to 0.1, and sliced regularizer with $\gamma$ equal to 1. We optimize for 5k steps using Adam \citep{Kingma2014} optimizer with 500 warm-up steps and effective batch size 65.5K tokens. \langpair{ro}{en} is fine-tune with learning rate $5\cdot10^{-5}$, \langpair{de}{en} model is fine-tuned with lr $7\cdot10^{-5}$. Details of the model size and hardware set up is shown in \Cref{tab:model-training-counts}

\begin{table}[]
    \centering
    \begin{tabular}{lcc}
    \toprule
       \multirow{2}{*}{} & \langpair{ro}{en} & \langpair{en}{de}\\ \midrule
        \# total & 46M & 356M \\
        \# trained & 4M & 51M\\
        time & 2h & 20h \\
        GPU & \multicolumn{2}{c}{NVIDIA GeForce GTX TITAN X} \\
        \# GPUs & \multicolumn{2}{c}{1} \\
        \bottomrule
    \end{tabular}
    \caption{\label{tab:model-training-counts}Total number of model parameters and number of trained parameters for \langpair{ro}{en} and \langpair{de}{en} alongside GPU model, number of GPUs and total training time.}
    \label{tab:placeholder}
\end{table}

\paragraph{Decoding Speed.} To make a fair comparison of the data store lookup, we use the same GPU cluster node with NVIDIA GeForce GTX TITAN X. 
\paragraph{\knnmt{} parameters.}
\knnmt{} relies on the three hyperparameters: number of retrieved neighbors ($k$), temperature ($T$), and interpolation parameter $\lambda$. \Cref{tab:knn-params-inf} shows the exact parameters we used for each \knnmt{} model.
\begin{table}[]
    \centering
    \begin{tabular}{lccc}
    \toprule
     test set &  $k$ & $T$ & $\lambda$  \\ \midrule
     \langpair{ro}{en} \texttt{newstest}& 8 & 100 & 0.3 \\
     \langpair{de}{en} \texttt{Law} & 8 & 10 & 0.8 \\
    \langpair{de}{en} \texttt{Medical} & 4 & 10 & 0.8 \\
    \langpair{de}{en} \texttt{IT} & 8 & 10 & 0.7\\
    \langpair{de}{en} \texttt{Koran} &8&100&0.7 \\
    \bottomrule
    \end{tabular}
    \caption{\label{tab:knn-params-inf} \knnmt{} parameters.}
    
\end{table}
\paragraph{Evaluation.}
We use SacreBLEU with the nrefs:1, case:mixed, eff:no, tok:13a, smooth:exp, version:2.3.1 
signature; and for COMET, we use default \texttt{Unbabel/wmt22-comet-da}\footnote{\url{https://huggingface.co/Unbabel/wmt22-comet-da}} model.

\section{Additional Results}
\label{app:add-results}
\Cref{tab:roen512} shows the results for \langpair{ro}{en} models with the output dimension of 512.
\begin{table}[ht]
    \centering
    \begin{tabular}{lccc}
    \toprule
    \textbf{model} & \textbf{BLEU}  & \textbf{COMET}   & tok./sec.\\
    \midrule
        base & 31.2 & 0.7855 & 94.6  \\ %
        base-D. & 31.1 & 0.7862 & 78.9 \\ \midrule %
        \knnmt{} & 31.8 & \textbf{0.7937} & 26.1 \\ %
        \knnmt{}-D & \textbf{31.9} & 0.7905 & 54.4 \\ %
    \bottomrule
    \end{tabular}
    \caption{\label{tab:roen512} BLEU and COMET on \texttt{newstest16} for \langpair{ro}{en} 512 model. \knnmt{} is with the in-domain data store.}
\end{table}

\section{Clustering Analysis \label{app:clustering_analysis}}
Here define homogeneity, completeness and $v$-measure clustering metrics \citep{rosenberg-hirschberg-2007-v}. To define these metrics, let us first denote the entropy and conditional entropy of clusters and next tokens: 
\begin{align*}
H(V|K) &= -\sum_{i=1}^{K} \sum_{v=1}^{|V|} \frac{n_{v,i}}{N} \log \frac{n_{v,i}}{\sum_{v=1}^{|V|} n_{v,i}} \\
H(V)   &= - \sum_{v=1}^{|V|} \frac{\sum_{i=1}^{K} n_{v,i}}{N} \log \frac{\sum_{i=1}^{K} n_{v,i}}{N} \\
H(K|V) &= -\sum_{v=1}^{|V|} \sum_{i=1}^{K} \frac{n_{v,i}}{N} \log \frac{n_{v,i}}{\sum_{i=1}^{K} n_{v,i}} \\
H(K)   &= - \sum_{i=1}^{K} \frac{\sum_{v=1}^{|V|} n_{v,i}}{N} \log \frac{\sum_{v=1}^{|V|} n_{v,i}}{N},
\end{align*}
where by $n_{v,i}$, we denote the number of data store elements with a next token $v$ and assigned with the cluster $i$.
Homogeneity measures if all of its clusters contain only data points which are members of a single class
\begin{equation}
    \text{hom} = \begin{cases}
      1 & \text{if}\, H(V)=1\\
      1 - \frac{H(V|K)}{H(V)} & \text{else};
    \end{cases}
\end{equation}
completeness measures if all the data points that are members of a given class are elements of the same cluster
\begin{equation}
    \text{compl} = \begin{cases}
      1 & \text{if}\, H(K)=1\\
      1 - \frac{H(K|V)}{H(K)} & \text{else};
    \end{cases}
\end{equation}
and $v$-measure is the harmonic mean between homogeneity and completeness, where $\beta$ weights the two components (we use $\beta=1$):
\begin{equation}
    v\text{-measure} = \frac{(1+\beta)\,\text{hom} \, \text{compl}}{(\beta \, \text{hom}) + \text{compl}}.
\end{equation}

\end{document}